\documentclass[letterpaper, 10 pt, conference]{ieeeconf}  

\IEEEoverridecommandlockouts                              
\usepackage{booktabs}
\usepackage{cite}
\usepackage{url}
\usepackage{graphicx}
\usepackage{epstopdf}
\usepackage{subfigure}
\usepackage{amssymb}
\usepackage{amsmath}
\usepackage{multirow}
\usepackage[colorlinks,
            linkcolor=red,
            anchorcolor=blue,
            citecolor=green
            ]{hyperref}

\title{\LARGE \bf
Object-to-Scene: Learning to Transfer Object Knowledge to Indoor Scene Recognition}

\author{Bo Miao$^{2,3}$, Liguang Zhou$^{1,2}$, Ajmal Saeed Mian$^{3}, \IEEEmembership{Senior Member,~IEEE}$, \\
Tin Lun Lam$^{1,2,\dagger}$, \IEEEmembership{Senior Member,~IEEE}, Yangsheng Xu$^{1,2}, \IEEEmembership{Fellow,~IEEE}$
\thanks{$^{\dagger}$Corresponding Author: Tin Lun Lam (tllam@cuhk.edu.cn)}
\thanks{This paper was supported in part by the funding AC01202101025 and 2019-INT007 from the Shenzhen Institute of Artificial Intelligence and Robotics for Society.}
\thanks{\textsuperscript{1} School of Science and Engineering, The Chinese University of Hong Kong, Shenzhen.}
\thanks{\textsuperscript{2} Shenzhen Institute of Artificial Intelligence and Robotics for Society, The Chinese University of Hong Kong, Shenzhen.}
\thanks{\textsuperscript{3} School of Physics, Mathematics and Computing, The University of Western Australia.}
}

\begin{document}

\maketitle
\thispagestyle{empty}
\pagestyle{empty}

\begin{abstract}
Accurate perception of the surrounding scene is helpful for robots to make reasonable judgments and behaviours. Therefore, developing effective scene representation and recognition methods are of significant importance in robotics. Currently, a large body of research focuses on developing novel auxiliary features and networks to improve indoor scene recognition ability. However, few of them focus on directly constructing object features and relations for indoor scene recognition. In this paper, we analyze the weaknesses of current methods and propose an Object-to-Scene (OTS) method, which extracts object features and learns object relations to recognize indoor scenes. The proposed OTS first extracts object features based on the segmentation network and the proposed object feature aggregation module (OFAM). Afterwards, the object relations are calculated and the scene representation is constructed based on the proposed object attention module (OAM) and global relation aggregation module (GRAM). The final results in this work show that OTS successfully extracts object features and learns object relations from the segmentation network. Moreover, OTS outperforms the state-of-the-art methods by more than 2\% on indoor scene recognition without using any additional streams. Code is publicly available at: \url{https://github.com/FreeformRobotics/OTS}.
\end{abstract}


\section{Introduction}
\label{Introduction}
Intelligent robots are human companions and assistants with the goal to improve the quality of life. For instance, robots can be designed to guide blind people to navigate by telling them where they are and which direction to go. For this application, there is an urgent need to equip robots with semantic scene understanding ability, e.g. the ability to perceive the surrounding objects and the scene that all objects constitute. Therefore, developing an effective method for indoor scene representation and recognition bears importance in improving the level of robotic perception and intelligence.

Indoor scene representation based recognition has been proposed for more than one decade, and it is still a challenging task in robotics and computer vision due to several issues: (1) conventional networks cannot focus on every object in scenes because they do not extract features at the object granularity \cite{He2016}; (2) handcrafted object features are sub-optimal to represent scenes \cite{Quelhas2005}; (3) an effective paradigm is lacking to represent coexisting object relations \cite{pal2019deduce}.

\begin{figure}[]
        \centering
        \includegraphics[width= 0.47\textwidth]{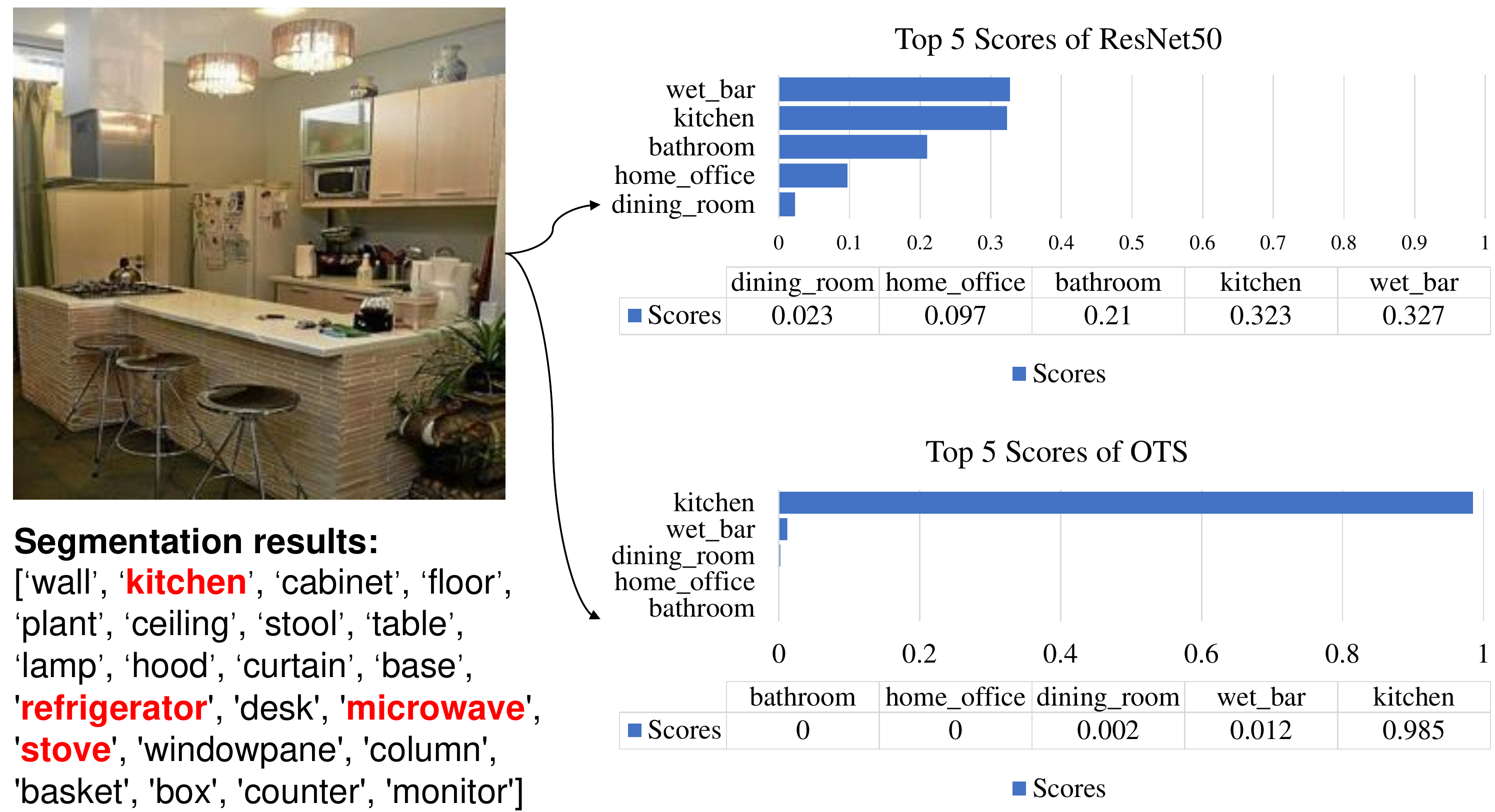}
        \vspace{-1mm}
        \caption{Comparison of the recognition results of the proposed OTS and ResNet50 on the shown image. The bottom left list shows the objects that the segmentation network in our method detected, and the right charts show the top 5 recognition scores of the two models.}
        \vspace{-5mm}

        \label{img1}
\end{figure}

This work primarily aims to mitigate the above-mentioned issues and improve the accuracy of indoor scene representation and recognition. To achieve these goals, we propose a novel one-stream method, namely Object-to-Scene (OTS), that constructs object features and calculates object relations to realize indoor scene representation and recognition. In this work, instead of simply using handcrafted features such as \cite{pal2019deduce} to represent the existence of each object, object features with higher semantic level are extracted using the proposed object feature aggregation module (OFAM) and a segmentation network. Object attention module (OAM) is then used to learn object relations implicitly, the attention mechanism of the module can obtain long-range dependencies, and thus coexisting objects in all situations can be learned. After that, Global relation aggregation module (GRAM) based on a strip depthwise convolution and a pointwise convolution is used to aggregate the object features and convert object features into a scene representation vector. Finally, a fully-connected layer is used for indoor scene recognition, and recognition results are compared with the existing state-of-the-art methods. Fig. \ref{img1} illustrates the superiority of our method, where the charts on the right show the top 5 recognition scores of our method and ResNet50 on the image. Compared to ResNet50 that focuses on major objects such as stool and table in scenes, the segmentation network in our method can detect tiny but important objects, such as stove and microwave in Fig. \ref{img1}, and our method can represent object features and relations implicitly. Therefore, ResNet50 hesitates between wet bar and kitchen but OTS, based on object features, can recognize kitchen confidently.

In summary, the major contributions of this paper are as follows:

\begin{itemize}
  \vspace{1mm}
  \item  We propose a novel framework OTS that enables using object features and relations for indoor scene representation and recognition, and OTS outperforms existing methods by more than 2\%.
  \vspace{1mm}
  \item We propose OFAM that can extract object features from the segmentation network for indoor scene representation and recognition.
  \vspace{1mm}
  \item We propose OAM to learn relations between objects, and it helps to improve the scene representation and recognition ability of our method. Meanwhile, the proposed object attention blocks in OAM are more flexible and effective compared with the well-known self-attention \cite{Zhang2019} and non-local \cite{Wang2017}.
  \vspace{1mm}
  \item We propose GRAM to fuse the object features and relations into scene representation at a higher semantic level.
\end{itemize}

The remainder of the paper is organized as follows. Section \ref{Related Works} describes a detailed list of recent works in scene representation and recognition. The details of the proposed OTS are described in Section \ref{Methodology}. Section \ref{Experiments and Results} presents the experimental settings and analyzes the results of OTS. Finally, the conclusion and future work are presented in Section \ref{Conclusions and Future Work}.

\vspace{2mm}
\section{Related Works}
\label{Related Works}

Understanding surrounding scenes is helpful for robots to make reasonable judgments and behaviours, and a proper scene representation is an important prerequisite \cite{Liao2016, Ye2017, Yan2019}. Many early approaches used statistics-based and hand-crafted features for scene representation \cite{Lazebnik2006, Khan2014}. Li et al. proposed codebooks to represent local features and scenes \cite{Fei-Fei2005}. Quattoni et al. used prototypes that contain object information for different indoor scene representations \cite{Quattoni2009}. Liu et al. proposed fast adaptive color tags to describe each indoor scene, and used tag matching methods in U-V color space and geometric space for inference \cite{Liu2009}. However, these statistics-based and hand-crafted features have limited semantic information and are sub-optimal to represent scenes as there are many combinations of objects in the same and different scenes. In that case, an effective scene representation method is urgently needed.

\begin{figure}[]
        \centering
        \includegraphics[width= 0.47\textwidth]{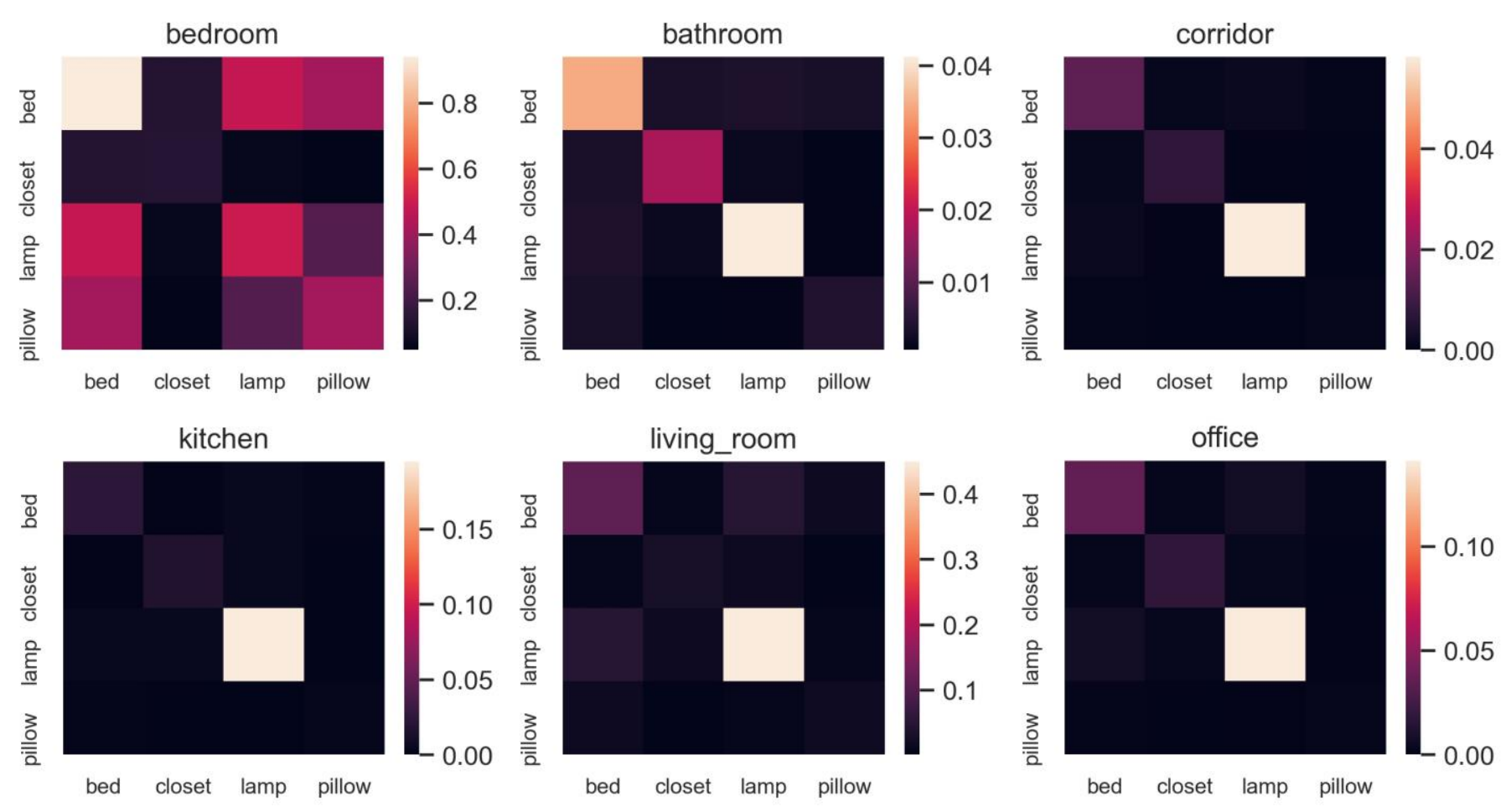}
        \vspace{-2mm}
        \caption{Object pair co-occurrence probability distribution of different scenes in Places365-7classes.}
        \vspace{-5mm}
        \label{img2}
\end{figure}

As computing power increases, many deep learning methods have been proposed for various vision tasks \cite{Lei2020,He2017mask,Hristov2020,Zhang2018,Herranz2016}. ResNet is one of the most prominent backbone feature extractors since it cleverly avoids the vanishing gradient problem in neural networks \cite{He2016}. However, ResNet behaves poorly in indoor scene representation and recognition due to the lack of effective expression of coexisting small objects and object relations. To solve this problem, some methods try to combine the backbone features with the object and semantic features from detection networks and segmentation networks for scene representation and recognition. Chen et al. merged backbone features, detection features and segmentation features into an embedding for indoor scene representation \cite{Chen2018}. Sun et al. used the method of spatial fisher vectors to extract object features from detection network, and combined the object features with contextual and global appearance features for scene recognition \cite{sun2018fusing}. Alejandro et al. combined the semantic features from segmentation network with backbone features for scene recognition using an attention module \cite{lopez2020semantic}. Pal et al. used detection network to compute a binary feature vector that represents whether each object existed in the scene, the binary object vector is then combined with backbone features for indoor scene recognition \cite{pal2019deduce}. Zhou et al. used a Bayesian method to represent the co-occurrence of object pairs for better scene representation and recognition \cite{Zhou21borm}. Zeng et al. added scene attributes into image features and patch features at multi-scale for scene recognition \cite{zeng2019}. Although the above mentioned methods improved the scene representation ability and recognition accuracy. These methods are sub-optimal in terms of constructing object features and relations. As shown in Fig. \ref{img2}, the co-occurrence probability of some coexisting objects, such as bed and lamp, varies significantly from the bedroom to other scenes. Therefore, learning object relations and object features well could further improve the representation ability and recognition accuracy of models. Inspired by the above observations, we propose a novel one-stream method called OTS that enables object features and relations for scene representation and recognition, and our method outperforms existing methods.

\vspace{-1mm}

\section{Methodology}
\label{Methodology}

\begin{figure}[t!]
    \centering
    \includegraphics[width= 0.48\textwidth]{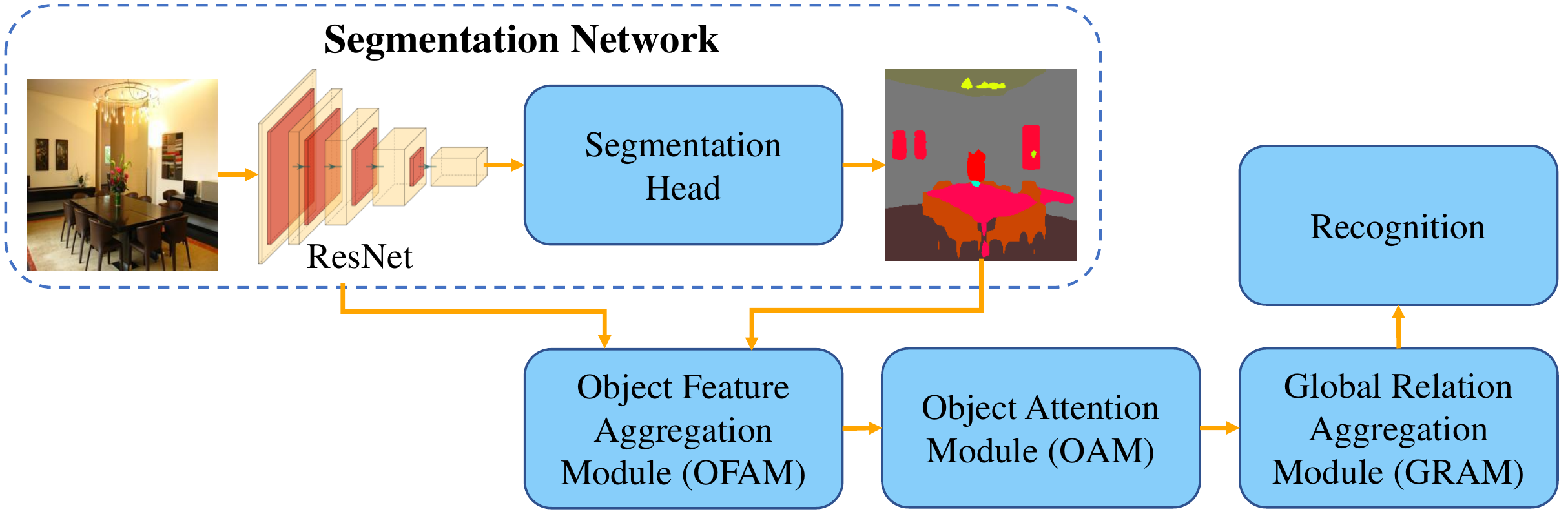}
    \caption{The proposed OTS includes five parts. (a) Segmentation network: PSPNet is used to calculate the segmentation mask of the input image and to provide its backbone features Conv4\_6 for the next step. (b) OFAM: the segmentation score map is combined with the Conv4\_6 feature map to form the object features $X_{obj}$. (c) OAM: cascaded object attention blocks in the OAM are used to calculate object relations. (d) GRAM: a large strip depthwise convolution with the size of the input feature map and a pointwise convolution are used to aggregate object features and form the final scene representation vector. (e) Recognition: a fully connected layer is used to recognize the scene.}
    \label{img3}
    \vspace{-4mm}
\end{figure}

We propose OTS for indoor scene representation and recognition. In OTS, we propose OFAM which enables small but discriminative objects to determine the final recognition results. To ensure effective utilization of object features for scene representation, we propose OAM to learn object relations based on its attention mechanism and a GRAM to fuse object features and relations. Fig. \ref{img3} illustrates the framework of the proposed OTS. Given an input image $I = (X_{i}, Y_{i})$, where $X_{i}$ is the image data and $Y_{i}$ is the corresponding category, OFAM is used to calculate the object features $X_{obj}$ at first. OAM and GRAM are then used to calculate the scene representation $V_{obj}$. Finally, a fully connected layer and the softmax function are used to calculate the final probability of each scene $P(Y|V_{obj})$.

\subsection{Object Feature Aggregation Module (OFAM)}
\label{OFAM}

\begin{figure}[t]
        \centering
        \includegraphics[width= 0.48\textwidth]{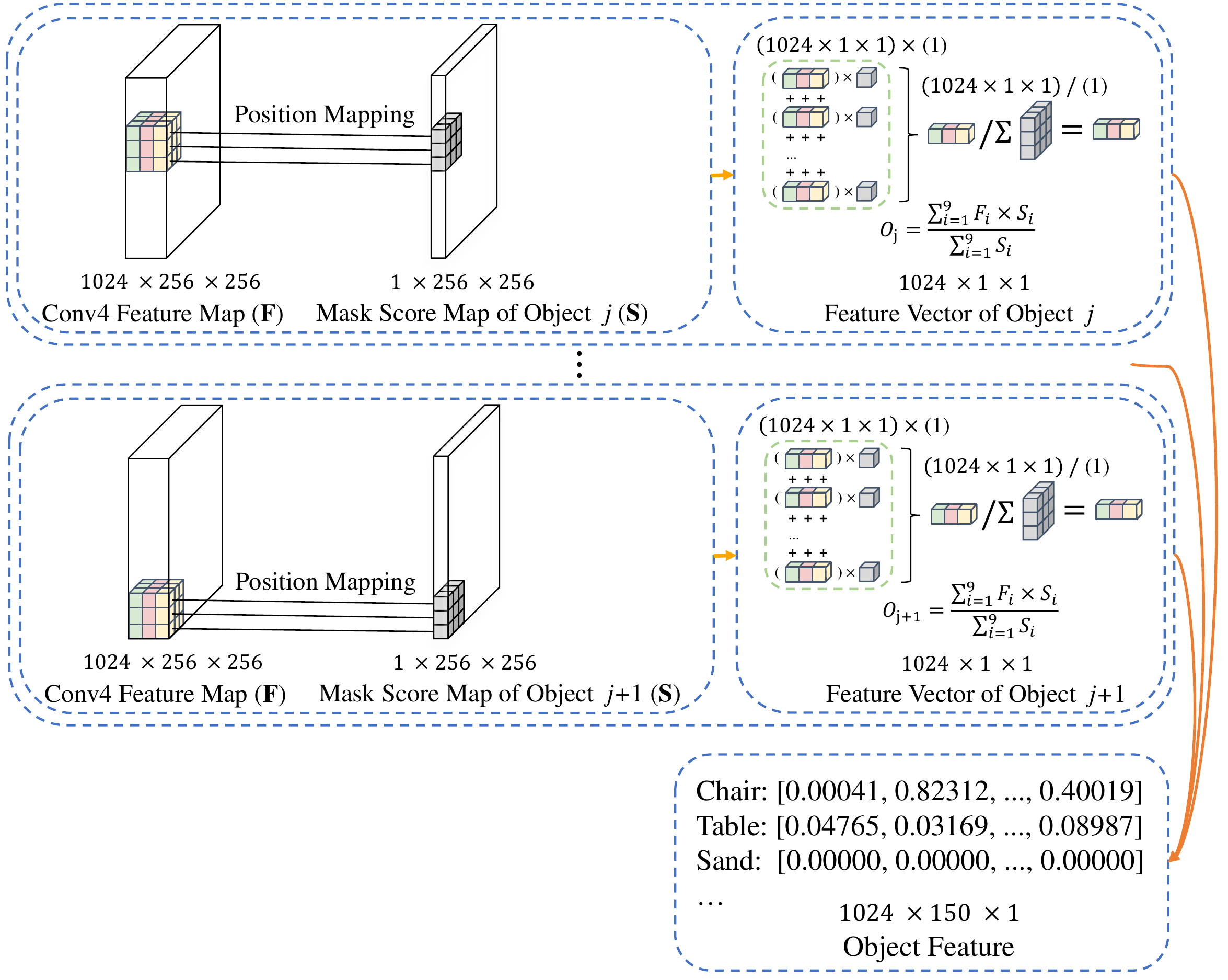}
        \vspace{-6mm}
        \caption{Framework of the proposed object feature aggregation module.}
        \label{img4}
        \vspace{-4mm}
\end{figure}

The proposed OFAM can extract object features based on the segmentation network. In this paper, PSPNet that is pretrained on ADE20k dataset is used as the segmentation network \cite{Zhao2017,Zhou2017}. Fig. \ref{img4} illustrates the proposed OFAM. OFAM uses Conv4\_6 feature map $F\in{R^{C\times N}}$ and segmentation score map $S\in{R^{C^{'}\times N}}$ that obtained from PSPNet to calculate object features $X_{obj}$, where $C$ is the number of channels, $C^{'}$ is the number of objects and $N$ is the number of feature spatial positions. In this paper, $C^{'}$ is set to 150 because of the object number of ADE20K, and $C$ is set to 1024 because Conv4\_6 is used. For convenience, we call each spatial position in $F$ as a unit, the channel values of each unit form the feature vector $U\in{R^{C\times 1}}$ of the unit. The binary mask $M\in{R^{C^{'}\times N}}$ of objects is calculated as:
\vspace{-1mm}
$$M_{i,j} = \begin{cases}
            1,\ if\ max(S_{i,1},\cdots,S_{i,C^{'}})\ == \ S_{i,j} \\
            0,\ otherwise.
            \end{cases} \eqno{(1)}$$
where $i$ is the unit index and $j$ is the object index. Then, the object feature vector $O_{j}\in{R^{C\times 1}}$ is calculated as:

$$O_{j} = \frac{\sum_{i=1}^{N}{(M_{i,j}*S_{i,j}*U_{i})}}{\sum_{i=1}^{N}{(M_{i,j}*S_{i,j})}} \eqno{(2)}$$
where $i$ is the unit index and $j$ is the object index, and $N$ represents the unit number. Finally, the feature vectors stacked together to form the object features $X_{obj}\in{R^{1024\times150}}$.

\subsection{Object Attention Module (OAM)}
\label{OAM}

In addition to object features, relations between objects are also required to improve scene representation ability and recognition accuracy. For example, a model will be more confident about the kitchen label if stove and refrigerator have been detected together. In that case, an attention mechanism is a good choice since it can capture long-range dependencies between all objects \cite{vaswani2017attention}. However, only global attention mechanism is suitable for our case because object features do not include spatial relations.

\begin{figure}[]
        \centering
        \includegraphics[width= 0.49\textwidth]{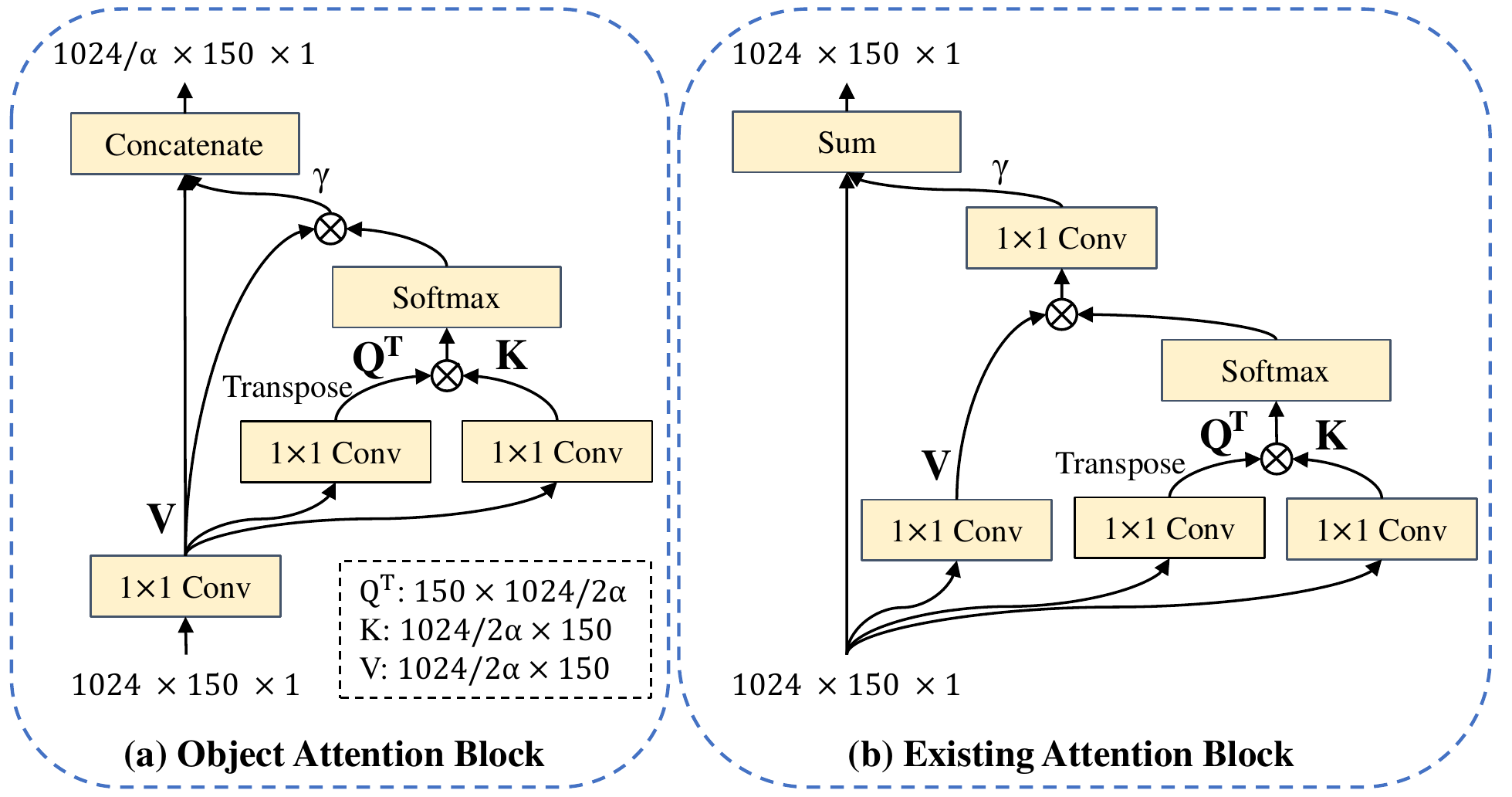}
        \vspace{-5mm}
        \caption{Structure of the proposed object attention block, non-local and self-attention.}
        \label{img5}
        \vspace{-5mm}
\end{figure}

Self-attention \cite{Zhang2019} and non-local \cite{Wang2017} are one of the most effective global attention mechanisms where three nodes are calculated in parallel with 1$\times$1 convolutions and the input feature map. We call the three nodes Query ($Q$), Key ($K$), and Value ($V$) for convenience. $Q$ and $K$ are then multiplied together to calculate the correlation matrix of each unit in the feature map, and softmax is used to calculate the activation map. After that, $V$ and the activation map are multiplied together, and an additional 1$\times$1 convolution is used to form the final attention. Finally, the attention multiplies a learnable scale parameter and adds back to the input feature map to form the output. Fig. \ref{img5}(b) illustrates the structure of self-attention and non-local. Although self-attention and non-local are effective, they can be further improved. For example, the attention will be more efficient if $Q$ and $K$ are calculated based on $V$, because $V$ has fewer channels compared with input features. Leveraging this insight, we propose an OAM that consists of cascaded object attention blocks as shown in Fig. \ref{img5}(a). Given an input feature map $F\in{R^{C\times N}}$, $V\in{R^{\frac{C}{2\alpha} \times N}}$ is first obtained using 1$\times$1 convolutions, where $C$ is the channel number, $N$ is the unit number, $\alpha$ is a factor to control the compression rate and the output channel number, and $V=W_{v}F$. $Q\in{R^{\frac{C}{2\alpha} \times N}}$ and $K\in{R^{\frac{C}{2\alpha} \times N}}$ are then calculated based on $V$, where $Q=W_{q}V$ and $K=W_{k}V$. The object attention is then calculated as follow:

\begin{tiny}
\begin{table}[b!]
\vspace{-3mm}
\tiny
\centering
\caption{Comparison of the proposed object attention block, non-local and self-attention.}\label{tab1}
\renewcommand
\setlength{\tabcolsep}{}{
\begin{tabular}{lllllllll}
\hline
&$C_{in}$&$C_{out}$&$C_{Q}$&$C_{K}$&$C_{V}$&Parm. (M)&FLOPs (M)\\
\hline
Self-attention&1024&1024&128&128&512&1.3&211.0\\
Non-local&1024&1024&512&512&512&2.1&337.6\\
Object Attention Block&1024&1024&512&512&512&\textbf{1.1}&\textbf{180.3}\\
\hline
\end{tabular}}
\end{table}
\end{tiny}

\vspace{-1mm}
$$\beta_{Q,K} = softmax(Q^{T}K)\eqno{(3)}$$
$$Attn = concatenate(\gamma*V\beta_{Q,K},\ V) \eqno{(4)}$$
Where $\gamma$ is a learnable scale parameter that can be updated gradually. The intuition for why we propose object attention block is straightforward. Firstly, $V$ is used to refine input information with fewer channels (512/$\alpha$) in object attention, calculating $Q$ and $K$ based on $V$ enables them to perceive refined input information with half the computational cost. Secondly, $\alpha$ is used to compress the block and control the output channel number, the connectivity of $Q$, $K$ and $V$ allows them to be automatically compressed proportionally with the change of $\alpha$. Thirdly, concatenation mechanism in object attention avoids using extra convolution and ensures the later block perceive the refined input information and attention features separately. Table \ref{tab1} compares the computational cost of three methods. It can be seen that our object attention block is the most efficient of the three methods under the same input and output channel number configuration. Our object attention block is also more flexible because we can adjust $\alpha$ to reduce the computational cost arbitrarily, and the channel number of $Q$, $K$, $V$, and output will be changed automatically.
\vspace{-1mm}

\subsection{Global Relation Aggregation Module (GRAM)}
\label{GRAM}

\begin{figure}[]
        \centering
        \includegraphics[width= 0.47\textwidth]{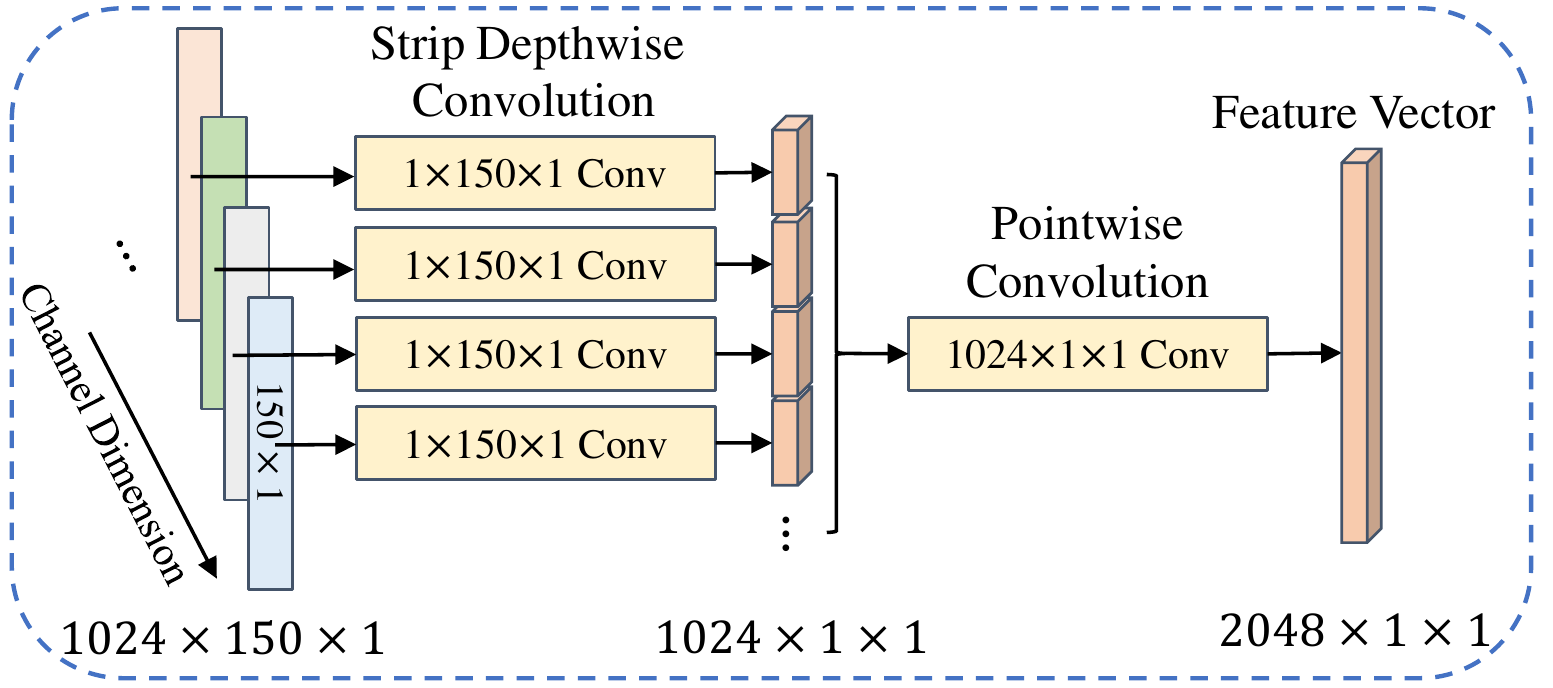}
        \vspace{-1mm}
        \caption{Structure of the proposed global relation aggregation module.}
        \label{img6}
        \vspace{-4mm}
\end{figure}

Aggregation of object features is essential for accurate scene recognition. To solve this problem, we propose a GRAM that consists of a strip depthwise convolution and a pointwise convolution as shown in Fig. \ref{img6}. Given an input feature map $F_{in}\in{R^{C\times N}}$, the strip depthwise convolution first aggregates object features and relations at each channel, and converts the feature map into $F_{mid}\in{R^{C\times 1}}$, pointwise convolution is then used to generate scene representation vector $F_{out}\in{R^{C^{'}\times 1}}$ that has higher semantic level. Depthwise convolution combined with pointwise convolution is much more efficient compared with conventional convolution kernel \cite{Francois2017}. The conventional depthwise convolution has a 3x3 kernel to learn local information in each channel. However, we aim to aggregate the object features and relations into a scene representation vector, and learn the global relation at the same time. Therefore, the conventional depthwise convolution is not suitable for our case. To solve this problem, we proposed a strip depthwise convolution, which operates on a list of object features rather than a patch of spatial features. The intuition for why we use this is clear. Firstly, the strip depthwise convolution can convert object feature map into a representation vector, that is more suitable for the final recognition layer. Secondly, the convolution is also able to aggregate the relations between all objects in each channel. Finally, the convolution is more efficient not only because of the depthwise characteristic, but also the avoiding of the use of large amount of fully connected layers.

\section{Experiments and Results}
\label{Experiments and Results}

\subsection{Implementation Details}
\label{Implementation Details}

Our network is implemented in the Pytorch library \cite{Paszke2019}, and a single RTX 2080Ti GPU is used in the experiments. We use the most common settings to implement the experiments \cite{He2016}. The batch size for all experiments is set to 256 and cross-entropy loss is used in our method. We adopt Stochastic Gradient Descent (SGD) optimizer with the base learning rate of 0.1 while momentum and weight decay are set to 0.9 and 0.0001, respectively. The learning rate is divided by 10 every 10 epochs. Training is performed for 40 epochs. Meanwhile, object features are calculated off-line to increase the training speed.

\subsection{Places365 Dataset}
\label{Places365 Dataset}

The reduced Places365 dataset is used in this paper since it is the largest scene recognition dataset with various indoor environment categories \cite{zhou2017places}. To verify the effectiveness of our method, we used two different class settings of indoor scenes for a fair comparison with other state-of-the-art methods. The first one contains 7 classes: Bathroom, Bedroom, Corridor, Dining room, Kitchen, Living room, and Office. We extract both training data and test data from the target 7 classes in Places365 and form the Places365-7classes, the class setting is totally the same as \cite{pal2019deduce}. The second one includes 14 indoor scenes in home environment: Balcony, Bedroom, Dining room, Home office, Kitchen, Living room, Staircase, Bathroom, Closet, Garage, Home theater, Laundromat, Playroom, and Wet bar. We also extract both training data and test data from the target 14 classes in Places365 and form the Places365-14classes, the class setting is totally the same as \cite{Chen2018}.

\subsection{SUN-RGBD Dataset}
\label{SUN-RGBD Dataset}

SUN-RGBD is one of the most challenging datasets for scene understanding \cite{song2015sun}, which includes images from various sources: 3784 images captured by Kinect v2; 1159 images captured by Intel RealSense cameras; 1449 images from NYU Depth V2 \cite{Silberman2012indoor} and 554 manually selected realistic scene images from Berkeley B3DO \cite{janoch2013category} captured by Kinect v1; 3389 selected frames by filtering out significantly blurred frames from the SUN3D videos \cite{xiao2013sun3d} captured by Asus Xtion. The diversity of categories and sources makes SUN-RGBD more suitable for verifying the generalization ability of methods. We only consider the RGB images in this work, and the 7 classes that chosen in Places365 are used to test our model's generalization ability. Notably, we only used the official test images in SUN-RGBD, and these images are evaluated by the model trained using Places365-7 classes.

\begin{table}[tp!]
\scriptsize
\centering
\caption{Scene recognition accuracy on the Places365-7classes.}\label{tab2}
\begin{tabular}{lllllll}
\hline
\multicolumn{1}{c}{} &
\multicolumn{2}{c}{\cite{He2016}} &
\multicolumn{3}{c}{\cite{pal2019deduce}} &
\multicolumn{1}{c}{Our} \\
\cmidrule(lr){2-3}\cmidrule(lr){4-6}\cmidrule(lr){7-7}
Category&ResNet18 &ResNet50 &Scene &Obj. &Scene+Obj. &OTS \\
\hline
Bathroom&87&94&92&65&91&92\\
Bedroom&82&83&90&74&90&97\\
Corridor&96&93&94&90&96&95\\
Dining room&81&71&79&94&79&88\\
Kitchen&83&84&87&62&87&92\\
Living room&55&66&84&25&80&79\\
Office&79&88&85&29&94&88\\
\hline
Avg. Acc. (\%)&80.4&82.7&87.3&62.6&88.1&\textbf{90.1}\\
\hline
\end{tabular}
\vspace{-6mm}
\end{table}

\begin{tiny}
\begin{table}[bp!]
\centering
\vspace{-3mm}
\caption{Scene recognition accuracy on the SUN-RGBD.}\label{tab3}
\begin{tabular}{ll|l|l|l}
\hline
Source&Method&1 Stream&2 Stream&Acc. (\%)\\
\hline
\multirow{2}{*}{\cite{He2016}} &ResNet18&\checkmark&& 63.3\\
&ResNet50&\checkmark&&67.2\\
\hline
\multirow{3}{*}{\cite{pal2019deduce}} &Scene&\checkmark&& 66.8\\
&Obj.&\checkmark&&53.6\\
&Scene+Obj.&&\checkmark&70.1\\
\hline
Our&OTS&\checkmark&&\textbf{70.6}\\
\hline
\end{tabular}
\end{table}
\end{tiny}

\begin{tiny}
\begin{table}[bp!]
\centering
\vspace{-1mm}
\caption{Scene recognition accuracy on the Places365-14classes.}\label{tab4}
\begin{tabular}{ll|l|l|l}
\hline
Source&Method&1 Stream&2 Stream&Acc. (\%)\\
\hline
\multirow{2}{*}{\cite{He2016}} &ResNet18&\checkmark&& 76.0\\
&ResNet50&\checkmark&& 80.0\\
\hline
\cite{Chen2018} & ResNet50+Word2Vec&&\checkmark& 83.7\\
\hline
Our & OTS &\checkmark&& \textbf{85.9}\\
\hline
\end{tabular}
\end{table}
\end{tiny}

\subsection{Main Results}
\label{Main Results}

To evaluate the effectiveness of the proposed OTS, we compare it with other benchmark methods on the Places365-7classes, Places365-14classes and reduced SUN-RGBD datasets. As shown in Table \ref{tab2}, OTS significantly outperforms the ResNet50 baseline with 7.4\% on Places365-7classes which shows the effectiveness of OTS. Moreover, OTS has 2\% higher accuracy than \cite{pal2019deduce} with only one stream. \cite{pal2019deduce} used a detection network to calculate one-hot object existence features, and added an additional stream to calculate auxiliary image features.

To verify the generalization ability of OTS, we further test OTS on SUN-RGBD. It is noteworthy that SUN-RGBD is just used as a test set to verify the generalization ability of models, and all models are trained using Plces365-7classes. As shown in Table \ref{tab3}, OTS achieves the highest accuracy compared with other methods on SUN-RGBD. The results demonstrate the generalization ability of OTS.

We further compare OTS with \cite{Chen2018} on the larger Places365-14classes. \cite{Chen2018} used a segmentation network to calculate Word2Vec features, and added an additional stream to calculate image features. As shown in Table \ref{tab4}, our OTS is 2.2\% higher than \cite{Chen2018} without any additional streams. The results show the effectiveness of our method. Better segmentation models have the potential to improve the performance of OTS, but they are not the focus of this work. In addition to evaluating the benchmark datasets, we also inference OTS in a real-world office environment, which can be found in our video supplement files.

\begin{tiny}
\begin{table}[tp!]
\centering
\caption{Ablation studies on each module.}\label{tab5}
\begin{tabular}{lllll}
\hline
ResNet50&OFAM&OAM&GRAM&Acc. (\%)\\
\hline
\checkmark&&&&80.0\\
\checkmark&\checkmark&&&77.2\\
\checkmark&\checkmark&\checkmark&&82.0\\
\checkmark&\checkmark&\checkmark&\checkmark&\textbf{85.9}\\
\hline
\end{tabular}
\vspace{-6mm}
\end{table}
\end{tiny}

\begin{figure*}[t!]
        \centering
        \includegraphics[width= 0.95\textwidth]{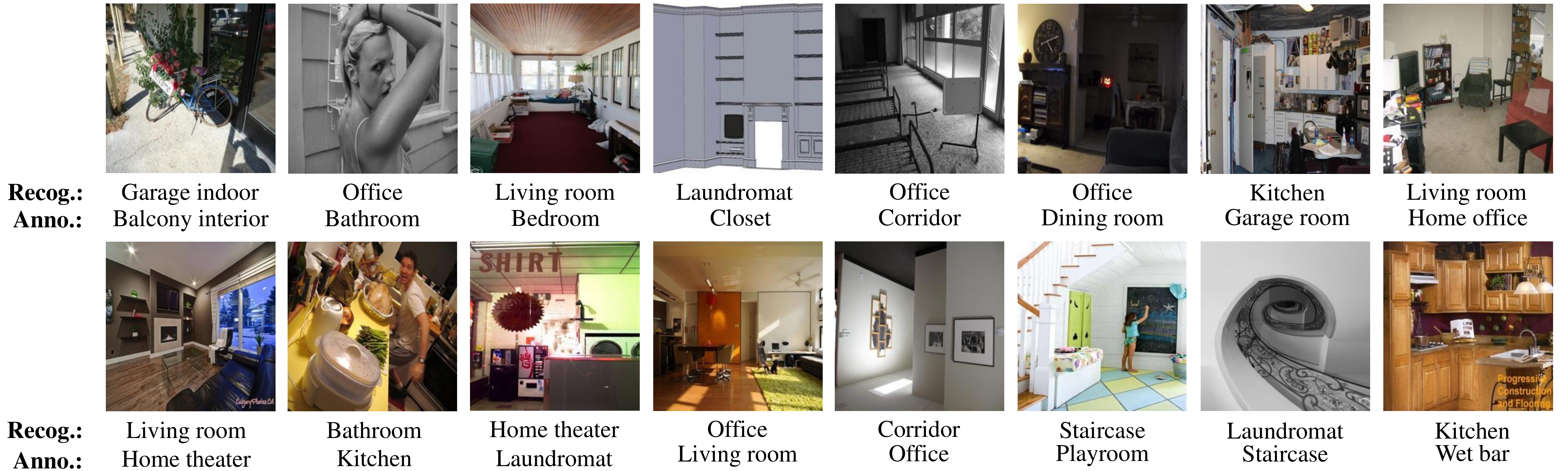}
        \vspace{-1mm}
        \caption{Failure cases of OTS. Recog. refers to the recognition results of OTS, and Anno. refers to the annotations.}
        \label{img7}
        \vspace{-4mm}
\end{figure*}

\subsection{Ablation Experiments}
\label{Ablation Experiments}

We run ablation experiments to analyze the results of our method. Unless specified otherwise, the dataset used in ablation experiments is Places365-14classes since it is larger and contains more categories. We firstly run an ablation experiment to show the effectiveness of each module. As shown in Table \ref{tab5}, both OAM and GRAM significantly improve the scene recognition accuracy because of the ability to capture the long-range dependencies between all objects. Meanwhile, only adding OFAM degrades the performance of the model because ResNet50 uses the features of Conv5 layer but OFAM extracts the object features from Conv4 layer as shown in Fig. \ref{img3}, which means the performance gains of our method rely on the object relation construction instead of the backbone features of the pre-trained segmentation model.

\begin{tiny}
\begin{table}[bp!]
\centering
\vspace{-3mm}
\renewcommand\tabcolsep{4pt}
\caption{Ablation studies on the combination methods in the proposed object attention block (OAB).}\label{tab6}
\begin{tabular}{lllllll}
\hline
&Num.&SUM.&CAT.&Acc. (\%)&Parm. (M)&FLOPs (M)\\
\hline
OAB&1&\checkmark&&85.0&0.4&70.5\\
OAB&2&\checkmark&&85.2&1.8&321.3\\
OAB&1&&\checkmark&\textbf{85.4}&\textbf{0.4}&\textbf{70.5}\\
OAB&2&&\checkmark&\textbf{85.9}&\textbf{1.2}&\textbf{211.5}\\
\hline
\end{tabular}
\end{table}
\end{tiny}

\vspace{1mm}
\subsubsection{Number of object attention blocks}
\label{Number of object attention blockss}

As stated in Section \ref{OAM}, OAM is built based on one or several cascaded object attention blocks. Therefore, we run ablation experiments to find how many object attention blocks is the optimal choice to balance efficiency and effectiveness. Unless specified otherwise, the $\alpha$ of the first object attention block is 2 and the $\alpha$ of the second object attention block is 0.5 to control the output channel number. Before that, we first verify the difference between using summation (SUM.) and using concatenation (CAT.) in the last step of object attention blocks. The experiment results are shown in Table \ref{tab6}. It can be seen that using concatenation in object attention blocks is both more efficient and effective compared with using summation. This is because simple summation cannot perfectly fuse object features and object relation features together, but concatenation can store both types of features separately. Then, we use concatenation in object attention blocks to find the optimal number of blocks. As shown in Table \ref{tab7}, we list the results as well as FLOPs obtained with different number of object attention blocks. It can be seen that when two object attention blocks are cascaded, OTS has the best accuracy 85.9\%. However, adding more object attention blocks exert an adverse impact on the results because these blocks make the model too complicated.

\begin{tiny}
\begin{table}[tp!]
\centering
\caption{Ablation studies on the number of the proposed object attention block (OAB).}\label{tab7}
\begin{tabular}{lllll}
\hline
&Num.&Acc. (\%)&Parm. (M)&FLOPs (M)\\
\hline
OAB&1&85.4&\textbf{0.4}&\textbf{70.5}\\
OAB&2&\textbf{85.9}&1.2&211.5\\
OAB&3&85.0&1.4&262.3\\
OAB&4&85.0&1.7&313.2\\
\hline
\end{tabular}
\vspace{-2mm}
\end{table}
\end{tiny}

\vspace{1mm}
\subsubsection{Object attention block vs Self-attention and Non-local}
\label{Object attention block vs Self-attention and Non-local}

We have shown the effectiveness of OAM in Table \ref{tab5}. In addition, We run an ablation experiment to show the effectiveness and efficiency of the proposed object attention blocks in OAM. As shown in Table \ref{tab8}, the accuracy of the proposed object attention block has an improvement of 1.2\% compared with self-attention, and an improvement of 3.6\% compared with non-local. What is more, the proposed object attention block has the least FLOPs compared with others under a similar condition. The object attention block (S) means only one object attention block is used in OAM.

\vspace{1mm}
\subsubsection{Global relation aggregation module (GRAM)}
\label{Global relation aggregation module (GRAM)}

We set two kinds of control groups to show the effectiveness of the proposed GRAM. In the ablation experiment, fully connected layer and Pooling layer are used to replace GRAM, separately. As shown in Table \ref{tab9}, GRAM obtained the best accuracy with only 2.3M parameters. It is noteworthy that FC in Table \ref{tab9} is equal to a large conventional convolution kernel with the size of input feature map. FC not only has more computational cost, but also has inferior performance compared with the proposed GRAM. The results indicate that GRAM can balance efficiency and effectiveness well.

\begin{tiny}
\begin{table}[tp!]
\centering
\caption{Ablation studies on the object attention block (OAB), non-local, and self-attention.}\label{tab8}
\begin{tabular}{llll}
\hline
&Acc. (\%)&Parm. (M)&FLOPs (M)\\
\hline
Non-Local \cite{Wang2017}&82.3&4.2&675.2\\
Self-Attention \cite{Zhang2019}&84.7&2.6&422.0\\
Object Attention Block&\textbf{85.9}&\textbf{1.2}&\textbf{211.5}\\
Object Attention Block (S)&\textbf{85.4}&\textbf{0.4}&\textbf{70.5}\\
\hline
\end{tabular}
\vspace{-5mm}
\end{table}
\end{tiny}

\begin{tiny}
\begin{table}[tp!]
\centering
\caption{Ablation studies on the GRAM.}\label{tab9}
\begin{tabular}{llll}
\hline
&Acc. (\%)&Parm. (M)&FLOPs (M)\\
\hline
FC&85.0&314.6&314.6\\
Max \& Avg. Pooling&82.0&\textbf{0}&\textbf{0.3}\\
GRAM&\textbf{85.9}&2.3&2.3\\
\hline
\end{tabular}
\vspace{-5mm}
\end{table}
\end{tiny}

\vspace{1mm}
\subsubsection{Failure cases}
\label{Failure cases}
During experiments, we find that ResNet50 misclassified many common images. However, OTS successfully recognized the common images in scenes because it can detect the objects and learn their relations in each scene. Then, we analyze the failure cases of OTS to demonstrate the pros and cons. As shown in Fig. \ref{img7}, OTS misclassified some images in each scene mainly caused by the missed or wrong detection. For example, kitchen is misclassified as bathroom because the pot is mistaking for a toilet, and office is misclassified as corridor because only wall, painting, ceiling, floor, and light are detected, which are the common coexisting objects in corridor. Therefore, the performance of OTS could be further improved with the help of more accurate segmentation results.

\section{Conclusions and Future Work}
\label{Conclusions and Future Work}

In this paper, we analyzed the weakness of existing scene representation and recognition methods, and proposed OTS to solve these issues. We further demonstrated that OTS can effectively use object features and relations for scene representation and recognition by comparing OTS with other existing state-of-the-art methods. Based on numerous ablation experiments, we also showed that OAM and GRAM perform well in learning object relations for scene representation. Moreover, the results of our work reflect several interesting conclusions: 1) object features can perform well as long as an appropriate object feature and relation learning method is used; 2) the backbone features in segmentation network can also be used for scene recognition instead of adding an additional stream to calculate; 3) attention mechanism is very suitable for computing object relations. We hope these results could guild future works for scene understanding. What is more, our work can also be promoted in the future. First of all, data augmentation methods can be used to extract more diverse object features during training. Secondly, enriching the number of object categories can offer a better scene representation, and thus improve scene recognition ability of the model. In the future, we plan to extend our methods to other mobile robots and establish more accurate semantic maps. Therefore, they can be better used to improve human's life quality.

\bibliographystyle{IEEEtran}
\bibliography{Ref}

\begin{thebibliography}{10}
\providecommand{\url}[1]{#1}
\csname url@samestyle\endcsname
\providecommand{\newblock}{\relax}
\providecommand{\bibinfo}[2]{#2}
\providecommand{\BIBentrySTDinterwordspacing}{\spaceskip=0pt\relax}
\providecommand{\BIBentryALTinterwordstretchfactor}{4}
\providecommand{\BIBentryALTinterwordspacing}{\spaceskip=\fontdimen2\font plus
\BIBentryALTinterwordstretchfactor\fontdimen3\font minus
  \fontdimen4\font\relax}
\providecommand{\BIBforeignlanguage}[2]{{%
\expandafter\ifx\csname l@#1\endcsname\relax
\typeout{** WARNING: IEEEtran.bst: No hyphenation pattern has been}%
\typeout{** loaded for the language `#1'. Using the pattern for}%
\typeout{** the default language instead.}%
\else
\language=\csname l@#1\endcsname
\fi
#2}}
\providecommand{\BIBdecl}{\relax}
\BIBdecl

\bibitem{He2016}
K.~He, X.~Zhang, S.~Ren, and J.~Sun, ``Deep residual learning for image
  recognition,'' in \emph{Proceedings of the IEEE conference on computer vision
  and pattern recognition}, 2016, pp. 770--778.

\bibitem{Quelhas2005}
P.~Quelhas, F.~Monay, J.-M. Odobez, D.~Gatica-Perez, T.~Tuytelaars, and
  L.~Van~Gool, ``Modeling scenes with local descriptors and latent aspects,''
  in \emph{Tenth IEEE International Conference on Computer Vision (ICCV'05)
  Volume 1}, vol.~1.\hskip 1em plus 0.5em minus 0.4em\relax IEEE, 2005, pp.
  883--890.

\bibitem{pal2019deduce}
A.~Pal, C.~Nieto-Granda, and H.~I. Christensen, ``Deduce: Diverse scene
  detection methods in unseen challenging environments,'' in \emph{2019
  IEEE/RSJ International Conference on Intelligent Robots and Systems (IROS)},
  2019, pp. 4198--4204.

\bibitem{Zhang2019}
H.~Zhang, I.~Goodfellow, D.~Metaxas, and A.~Odena, ``Self-attention generative
  adversarial networks,'' in \emph{International Conference on Machine
  Learning}, 2019, pp. 7354--7363.

\bibitem{Wang2017}
X.~Wang, R.~Girshick, A.~Gupta, and K.~He, ``Non-local neural networks,'' in
  \emph{Proceedings of the IEEE conference on computer vision and pattern
  recognition}, 2017, pp. 7794--7803.

\bibitem{Liao2016}
Y.~Liao, S.~Kodagoda, Y.~Wang, L.~Shi, and Y.~Liu, ``Understand scene
  categories by objects: A semantic regularized scene classifier using
  convolutional neural networks,'' in \emph{2016 IEEE international conference
  on robotics and automation (ICRA)}.\hskip 1em plus 0.5em minus 0.4em\relax
  IEEE, 2016, pp. 2318--2325.

\bibitem{Ye2017}
C.~Ye, Y.~Yang, R.~Mao, C.~Ferm{\"u}ller, and Y.~Aloimonos, ``What can i do
  around here? deep functional scene understanding for cognitive robots,'' in
  \emph{2017 IEEE International Conference on Robotics and Automation
  (ICRA)}.\hskip 1em plus 0.5em minus 0.4em\relax IEEE, 2017, pp. 4604--4611.

\bibitem{Yan2019}
F.~Yan, S.~Nannapaneni, and H.~He, ``Robotic scene understanding by using a
  dictionary,'' in \emph{2019 IEEE International Conference on Robotics and
  Biomimetics (ROBIO)}.\hskip 1em plus 0.5em minus 0.4em\relax IEEE, 2019, pp.
  895--900.

\bibitem{Lazebnik2006}
S.~Lazebnik, C.~Schmid, and J.~Ponce, ``Beyond bags of features: Spatial
  pyramid matching for recognizing natural scene categories,'' in \emph{2006
  IEEE Computer Society Conference on Computer Vision and Pattern Recognition
  (CVPR'06)}, vol.~2.\hskip 1em plus 0.5em minus 0.4em\relax IEEE, 2006, pp.
  2169--2178.

\bibitem{Khan2014}
S.~H. Khan, M.~Bennamoun, F.~Sohel, and R.~Togneri, ``Geometry driven semantic
  labeling of indoor scenes,'' in \emph{European Conference on Computer
  Vision}.\hskip 1em plus 0.5em minus 0.4em\relax Springer, 2014, pp. 679--694.

\bibitem{Fei-Fei2005}
L.~Fei-Fei and P.~Perona, ``A bayesian hierarchical model for learning natural
  scene categories,'' in \emph{2005 IEEE Computer Society Conference on
  Computer Vision and Pattern Recognition (CVPR'05)}, vol.~2.\hskip 1em plus
  0.5em minus 0.4em\relax IEEE, 2005, pp. 524--531.

\bibitem{Quattoni2009}
A.~Quattoni and A.~Torralba, ``Recognizing indoor scenes,'' in \emph{2009 IEEE
  Conference on Computer Vision and Pattern Recognition}.\hskip 1em plus 0.5em
  minus 0.4em\relax IEEE, 2009, pp. 413--420.

\bibitem{Liu2009}
M.~Liu, D.~Scaramuzza, C.~Pradalier, R.~Siegwart, and Q.~Chen, ``Scene
  recognition with omnidirectional vision for topological map using lightweight
  adaptive descriptors,'' in \emph{2009 IEEE/RSJ International Conference on
  Intelligent Robots and Systems}, 2009, pp. 116--121.

\bibitem{Lei2020}
H.~Lei, N.~Akhtar, and A.~Mian, ``Spherical kernel for efficient graph
  convolution on 3d point clouds,'' \emph{IEEE Transactions on Pattern Analysis
  and Machine Intelligence}, 2020.

\bibitem{He2017mask}
K.~He, G.~Gkioxari, P.~Doll{\'a}r, and R.~Girshick, ``Mask r-cnn,'' in
  \emph{Proceedings of the IEEE international conference on computer vision},
  2017, pp. 2961--2969.

\bibitem{Hristov2020}
Y.~Hristov, D.~Angelov, M.~Burke, A.~Lascarides, and S.~Ramamoorthy,
  ``Disentangled relational representations for explaining and learning from
  demonstration,'' in \emph{Conference on Robot Learning}, 2020, pp. 870--884.

\bibitem{Zhang2018}
L.~Zhang, G.~Zhu, L.~Mei, P.~Shen, S.~A.~A. Shah, and M.~Bennamoun, ``Attention
  in convolutional lstm for gesture recognition,'' in \emph{Advances in Neural
  Information Processing Systems}, 2018, pp. 1953--1962.

\bibitem{Herranz2016}
L.~Herranz, S.~Jiang, and X.~Li, ``Scene recognition with cnns: objects, scales
  and dataset bias,'' in \emph{Proceedings of the IEEE Conference on Computer
  Vision and Pattern Recognition}, 2016, pp. 571--579.

\bibitem{Chen2018}
B.~X. Chen, R.~Sahdev, D.~Wu, X.~Zhao, M.~Papagelis, and J.~K. Tsotsos, ``Scene
  classification in indoor environments for robots using context based word
  embeddings,'' in \emph{2018 IEEE International Conference on Robotics and
  Automation (ICRA) Workshop}, 2018.

\bibitem{sun2018fusing}
N.~Sun, W.~Li, J.~Liu, G.~Han, and C.~Wu, ``Fusing object semantics and deep
  appearance features for scene recognition,'' \emph{IEEE Transactions on
  Circuits and Systems for Video Technology}, vol.~29, no.~6, pp. 1715--1728,
  2018.

\bibitem{lopez2020semantic}
A.~L{\'o}pez-Cifuentes, M.~Escudero-Vi{\~n}olo, J.~Besc{\'o}s, and
  {\'A}.~Garc{\'\i}a-Mart{\'\i}n, ``Semantic-aware scene recognition,''
  \emph{Pattern Recognition}, vol. 102, p. 107256, 2020.

\bibitem{Zhou21borm}
L.~Zhou, C.~Jun, X.~Wang, Z.~Sun, T.~L. Lam, and Y.~Xu, ``Borm: Bayesian object
  relation model for indoor scene recognition,'' in \emph{IEEE/RSJ
  International Conference on Intelligent Robots and Systems}.\hskip 1em plus
  0.5em minus 0.4em\relax IEEE, 2021.

\bibitem{zeng2019}
H.~Zeng, X.~Song, G.~Chen, and S.~Jiang, ``Learning scene attribute for scene
  recognition,'' \emph{IEEE Transactions on Multimedia}, 2019.

\bibitem{Zhao2017}
H.~Zhao, J.~Shi, X.~Qi, X.~Wang, and J.~Jia, ``Pyramid scene parsing network,''
  in \emph{Proceedings of the IEEE conference on computer vision and pattern
  recognition}, 2017, pp. 2881--2890.

\bibitem{Zhou2017}
B.~Zhou, H.~Zhao, X.~Puig, S.~Fidler, A.~Barriuso, and A.~Torralba, ``Scene
  parsing through ade20k dataset,'' in \emph{Proceedings of the IEEE conference
  on computer vision and pattern recognition}, 2017, pp. 633--641.

\bibitem{vaswani2017attention}
A.~Vaswani, N.~Shazeer, N.~Parmar, J.~Uszkoreit, L.~Jones, A.~N. Gomez,
  {\L}.~Kaiser, and I.~Polosukhin, ``Attention is all you need,'' in
  \emph{Advances in neural information processing systems}, 2017, pp.
  5998--6008.

\bibitem{Francois2017}
C.~Fran{\c{c}}ois, ``Xception: Deep learning with depthwise separable
  convolutions,'' in \emph{Proceedings of the IEEE conference on computer
  vision and pattern recognition}, 2017, pp. 1800--1807.

\bibitem{Paszke2019}
A.~Paszke, S.~Gross, F.~Massa, A.~Lerer, J.~Bradbury, G.~Chanan, T.~Killeen,
  Z.~Lin, N.~Gimelshein, L.~Antiga \emph{et~al.}, ``Pytorch: An imperative
  style, high-performance deep learning library,'' in \emph{Advances in neural
  information processing systems}, 2019, pp. 8026--8037.

\bibitem{zhou2017places}
B.~Zhou, A.~Lapedriza, A.~Khosla, A.~Oliva, and A.~Torralba, ``Places: A 10
  million image database for scene recognition,'' \emph{IEEE Transactions on
  Pattern Analysis and Machine Intelligence}, 2017.

\bibitem{song2015sun}
S.~Song, S.~P. Lichtenberg, and J.~Xiao, ``Sun rgb-d: A rgb-d scene
  understanding benchmark suite,'' in \emph{Proceedings of the IEEE conference
  on computer vision and pattern recognition}, 2015, pp. 567--576.

\bibitem{Silberman2012indoor}
N.~Silberman, D.~Hoiem, P.~Kohli, and R.~Fergus, ``Indoor segmentation and
  support inference from rgbd images,'' in \emph{European conference on
  computer vision}.\hskip 1em plus 0.5em minus 0.4em\relax Springer, 2012, pp.
  746--760.

\bibitem{janoch2013category}
A.~Janoch, S.~Karayev, Y.~Jia, J.~T. Barron, M.~Fritz, K.~Saenko, and
  T.~Darrell, ``A category-level 3d object dataset: Putting the kinect to
  work,'' in \emph{Consumer depth cameras for computer vision}.\hskip 1em plus
  0.5em minus 0.4em\relax Springer, 2013, pp. 141--165.

\bibitem{xiao2013sun3d}
J.~Xiao, A.~Owens, and A.~Torralba, ``Sun3d: A database of big spaces
  reconstructed using sfm and object labels,'' in \emph{Proceedings of the IEEE
  International Conference on Computer Vision}, 2013, pp. 1625--1632.

\end{thebibliography}


\end{document}